\documentclass[10pt,twocolumn,letterpaper]{article}

\usepackage{wacv}              
\usepackage{amsmath}
\usepackage{graphicx}
\usepackage{amssymb}
\usepackage{booktabs}
\usepackage[utf8]{inputenc} 
\usepackage[T1]{fontenc}    
\usepackage{url}            
\usepackage{booktabs}       
\usepackage{amsfonts}       
\usepackage{nicefrac}       
\usepackage{microtype}      
\usepackage{xcolor}         
\usepackage{cite}
\usepackage{graphicx}
\usepackage{amsmath}
\usepackage{algorithm}
\usepackage{algpseudocode} 
\usepackage{tabularx}
\usepackage{wrapfig}
\usepackage{pifont}
\usepackage{placeins}
\usepackage{siunitx}
\usepackage{makecell}
\usepackage[accsupp]{axessibility}
\usepackage[pagebackref,breaklinks,colorlinks]{hyperref}

\usepackage[capitalize]{cleveref}
\crefname{section}{Sec.}{Secs.}
\Crefname{section}{Section}{Sections}
\Crefname{table}{Table}{Tables}
\crefname{table}{Tab.}{Tabs.}

\def\wacvPaperID{2293} 
\def\confName{WACV}
\def\confYear{2025}

\begin{document}

\title{FAIR-TAT: Improving Model Fairness Using Targeted Adversarial Training}

\author{Tejaswini Medi$^{1}$, Steffen Jung$^{1,2}$, Margret Keuper$^{1,2}$ \\
\small $^1$University of Mannheim, Germany $^2$MPI for Informatics, Saarland Informatics Campus, Germany \\
\small tejaswini.medi@uni-mannheim.de
}

\maketitle

\begin{abstract}
   Deep neural networks are susceptible to adversarial attacks and common corruptions, which undermine their robustness.
   In order to enhance model resilience against such challenges, Adversarial Training (AT) has emerged as a prominent solution.
   Nevertheless, adversarial robustness is often attained at the expense of model fairness during AT, i.e., disparity in class-wise robustness of the model.
   While distinctive classes become more robust towards such adversaries, hard to detect classes suffer.
   Recently, research has focused on improving model fairness specifically for perturbed images, overlooking the accuracy of the most likely non-perturbed data.
   Additionally, despite their robustness against the adversaries encountered during model training, state-of-the-art adversarial trained models have difficulty maintaining robustness and fairness when confronted with diverse adversarial threats or common corruptions.
   In this work, we address the above concerns by introducing a novel approach called Fair Targeted Adversarial Training (FAIR-TAT).
   We show that using targeted adversarial attacks for adversarial training (instead of untargeted attacks) can allow for more favorable trade-offs with respect to adversarial fairness.
   Empirical results validate the efficacy of our approach.
\end{abstract}

\section{Introduction}
\label{sec:intro}
Deep neural networks have led to tremendous success in a wide range of computer vision applications \cite{he_deep_2016, girshick2015fast, kirillov2023segment}.
One of the most relevant remaining challenges is their lack of robustness towards small domain shifts~\cite{michaelis_benchmarking_2019}, common corruptions~\cite{hendrycks2019robustness,kar_3d_2022} and adversarially perturbed examples~\cite{goodfellow2015explaining,pgd,croce2021mind, agnihotri2023cospgd}.
This limits their usage in safety-critical applications 
\cite{ma2021understanding}.
Adversarial examples are artificially crafted samples, which do not alter the semantic meaning of natural samples.
However, they add a small amount of noise that alters the model's prediction.
As a prominent defense mechanism to such adversaries, models are trained using adversarial training (AT) algorithms \cite{zhao2023improving, madry2018towards, pmlr-v97-wang19i, zhang2019theoretically}, where adversarial samples along with clean samples are utilized during model training to build a more robust classifier. 
The most commonly used adversarial examples during AT are generated through Fast Gradient Sign Method (FGSM) \cite{goodfellow2015explaining} and Projected Gradient Descent (PGD) \cite{madry2018towards}.
These methods are overall efficient to compute and have shown to improve the overall model robustness \cite{pmlr-v97-wang19i, madry2018towards, zhang2019theoretically}.

Despite improving the overall model robustness, adversarial training has side effects: it usually reduces model accuracy on clean samples, and,  although accuracy on perturbed samples improves, there remains a noticeable accuracy gap between clean and perturbed samples \cite{tsipras2018robustness,zhang2019theoretically, wang2023generalist, yang2023improving}. 
A further significant observation is the impact of AT on model fairness \cite{wei2023cfa,xu2021robust,Benz2020RobustnessMB}, where a lack of fairness is measured  
by the disparity in class-wise performance on clean data and under attack.
Specifically, adversarially trained models tend to exhibit stronger robustness on distinctive classes but weak performance on harder to predict classes.
Thus, classes that are difficult to classify in clean data are even more susceptible to exploitation by attackers.
To avoid class-wise safety issues and a false sense of security for models with high average robustness, there is a strong 
need to mitigate the model discrimination against particular groups of classes and focus on improving performance on harder classes. 
To provide a simple, not politically motivated example, consider a model that tends to confuse dark feline, say black panthers, with brown bears but is good at distinguishing polar bears and white tigers.
All classes will benefit from AT to some extent, when attacked.
Yet, the benefit is small for black panthers and brown bears while it is large for tigers and polar bears.
At the same time, the clean accuracy will drop for all classes, even for those that benefit from AT on perturbed data.

Some works try to address class-wise biases and robustness differences by imposing fairness constraints during AT.
This can involve adjusting the weights of easy and hard classes \cite{xu2021robust} or changing the perturbation margin of attacks based on the class-wise performance of the model during AT \cite{xu2021robust, wei2023cfa}.
Recent works on model fairness \cite{ma2022tradeoff, sun2023improving, yue2024revisiting, xu2021robust, zhao2023improving, li2023wat, wang2023better} focus on improving the robustness of the worst-performing classes while preserving overall model robustness.
These works typically train using \textit{untargeted} adversaries, i.e.~adversaries that maximize the model's loss irrespective of the new predicted label (target class). 
In this work, we shift our focus on vulnerable classes (hard classes) and these classes are targeted more often than the stronger ones (easy classes) by generating more \textit{targeted} adversaries towards hard classes.
This approach helps to defend against perturbations targeting hard classes, 
and therefore reduces existing class-wise biases.

Our proposed AT framework is called Fair Targeted Adversarial Training (FAIR-TAT).
Unlike conventional AT, which focuses on training with \textit{untargeted} adversaries, FAIR-TAT focuses on \textit{targeted} adversaries.
This helps to better understand and steer the class-to-class 
biases by adjusting the training objectives of each class during AT.
We first examine class-to-class biases using standard metrics like "robust accuracy" and "class-wise false positive scores".
Based on the analysis using these metrics, we introduce a method that dynamically adjusts the adversarial training configuration during training.
Specifically, we sample the target classes for generating targeted adversaries during training by leveraging the dynamically monitored biases. 
As a result, models trained using FAIR-TAT show improved accuracy on hard classes for clean data as well as on perturbed data, i.e.~their model fairness is improved when compared to classical AT.
This finding also extends from adversarial attacks to common corruptions \cite{hendrycks2019robustness}. 
\\
The contributions of our work can be summarized as follows:

 \begin{itemize}
     \item We empirically examine class-specific vulnerabilities to misclassifications by analyzing class-wise false positives in adversarially trained models.
     Our findings show that these false positives highlight the most confused classes during training.
     Based on this, we prioritize these confused classes through generating targeted adversaries of these classes more often during AT and thus learn the class-specific biases. 
     \item We propose a framework called Fair Targeted Adversarial Training (FAIR-TAT), which uses targeted adversaries and adapts the training setup based on class-specific characteristics.
     Target class selection is guided by the distribution of class-wise false positives seen during instance-wise training.
     This approach ensures that samples are perturbed to resemble confused classes more frequently, improving class-wise accuracy balance at inference.
     \item FAIR-TAT can be combined with existing methods to enhance class-wise fairness.
     We show that FAIR-TAT surpasses state-of-the-art methods in fairness, not only against its trained adversaries but also against other adversaries and common corruptions \cite{hendrycks2019robustness}, while maintaining overall robustness. 
 \end{itemize}

\section{Related Work}
\paragraph{Adversarial Robustness.}
Adversarial robustness is a crucial concept for assessing the resilience of deep neural networks against adversarial attacks~\cite{goodfellow2015explaining,pgd,croce2021mind,agnihotri2023cospgd}. Significant progress has been made in evaluating robustness across diverse adversarially trained model architectures thoroughly~\cite{grabinski2022robust, croce2021robustbench}, while some studies have aimed to unify research efforts on architectural design choices and their influence on robustness~\cite{Jung2023}. Furthermore, advancements in adversarial robustness have been achieved across tasks such as image classification, restoration, and segmentation by enhancing or redesigning robust architectures~\cite{grabinski2024large,10.1007/978-3-031-73636-0_21,agnihotri2023unreasonable,grabinski2022aliasing, lukasik2023improving, Jung2023}. Despite these efforts, which focus on improving overall robust accuracy, consistent variations in robustness across specific classes have been observed. These findings highlight the nuanced challenges in achieving uniformly robust performance across all categories.

\paragraph{Adversarial Training.}
Models that are not trained with adversarial defenses are typically only robust to low-budget attacks, if at all. Adversarial training (AT) \cite{Yuan2017AdversarialEA, madry2018towards, pmlr-v80-athalye18a, goodfellow2015explaining} is a remedy to this problem, as it trains the model on adversarial samples optimized during training, effectively making out-of-domain attacks become in-domain samples. However, this approach can result in over-fitting to attacks used during training. Early stopping \cite{rice2020overfitting} and the addition of external (synthetic) data \cite{rebuffi2021data,gowal2021improving,wang2023better} have been proposed as effective solutions to address this problem. Prior works have analyzed AT from different perspectives including robust optimization~\cite{pmlr-v97-wang19i}, robust generalization~\cite{raghunathan2019adversarial}, and diverse training strategies~\cite{pmlr-v97-zhang19p,pang2020boosting,pmlr-v97-wang19i}.

\paragraph{Adversarial Fairness Problem.} Adversarial Fairness in machine learning is a growing concern, as models often favor specific groups of classes while underperforming on others after being trained adversarially.
This discrimination undermines the reliability of machine learning models.
Recent studies have investigated class biases in terms of model accuracy and robustness, advocating for fair training processes that classify all classes equally well~\cite{wei2023cfa, DBLP:conf/kdd/0001KJW021, xu2021robust, medi2024classwiserobustnessanalysis}.
These works also show that while AT can improve overall robustness, it often ignores discrepancies in the robustness of individual classes, amplifying unbalance in class-wise robustness.
Data distribution and adversarial learning algorithms can contribute to these class-wise discrimination problems, making hard classes more vulnerable to attacks.

To address unfairness, several methods have emerged.
Fair Robust Learning (FRL) \cite{xu2021robust} adjusts perturbation margins and class weights based on whether fairness constraints are violated, though this approach decreases overall accuracy.
Balanced Adversarial Training (BAT) \cite{sun2023improving} aims to balance fairness between source and target classes.
Class-wise Calibrated Fair Adversarial Training (CFA) \cite{wei2023cfa} addresses fairness by dynamically customizing adversarial configurations for different classes and modifying weight averaging techniques.
All these methods reweigh classes based on their vulnerability in terms of robustness during the optimization process.
Worst Class Adversarial Training (WAT) \cite{li2023wat} uses a no-regret dynamic algorithm to improve worst-class accuracy.
More recent work, DAFA \cite{lee2024dafa}, considers class similarity and assigns distinct adversarial margins and loss weights to each class, adjusting them to encourage a trade-off in robustness among similar classes. 

However, there exist classes which exhibit higher robustness performance but also frequently tend to be confused with other classes.
Therefore, in this work, we focus on customizing the training configuration of Adversarial Training (AT) based on class-wise confusion rates, which are evaluated using class-wise false positive scores. 



\section{Preliminaries}
\label{bias}
AT is defined as min-max optimization problem.
Given a model $f$ with parameters $\theta$, loss function $L$, and training data distribution $D$, the training algorithm tries to minimize the loss where the adversary aims to maximize the loss within a neighborhood, where the perturbation bound is ${\mathcal{B}(x, \epsilon)}$ = ${{x^\prime} : \lVert {x^\prime}-x \rVert_p \leq \epsilon}$ and $\epsilon$ is the perturbation margin.
\autoref{eq:eq_1} formalizes the AT objective as a min-max optimization problem, where ${x^\prime}$ represents the perturbed samples and $x$ represents the clean or natural samples.
%
\begin{align}
\label{eq:eq_1}
\min_{\theta} \mathbb{E}_{(x, y) \in D} \left[ \max_{ \lVert x^{\prime}-x \rVert_p \leq \epsilon } L(f_{\theta}(x^\prime), y) \right]
\end{align}

White-box attacks, i.e.~attacks with full access to the model and its weights, allow to efficiently find adversaries and are therefore commonly used for AT.  FGSM~\cite{goodfellow2015explaining} creates adversarial examples in a single step by updating the randomly initialized image perturbation in the direction of the maximal loss increase.
Specifically, it computes for an input image $x$ and its label $y$ an adversary $x^{\prime}$ according to \autoref{eq:eq_2}, where small values of $\epsilon$ ensure that perturbations are imperceptibly small:
  \begin{align}
\label{eq:eq_2}
    x^{\prime} = x +  \epsilon \cdot \text{sign}\left( \nabla_{x} L(f_{\theta}(x),y)\right).
\end{align}  

PGD \cite{madry2018towards} finds adversaries within the perturbation bound solving the inner maximization problem defined in \autoref{eq:eq_1}.
It employs a multi-step scheme to maximize the inner part of the saddle point formulation.
\autoref{eq_3} defines the PGD (untargeted) attack, where $\Pi$ represents the projection function, and $\alpha$ controls the gradient ascent step size:
\begin{align}
\label{eq_3}
x^{t+1} = \Pi_{\mathcal{B}(x, \epsilon)}(x^{t} + \alpha \cdot \text{sign}(\nabla_{x^t} L(f_{\theta}(x^t),y))).
\end{align}

 Different from the above adversarial training works, where untargeted white-box attacks are used to find the adversaries, we focus on \textit{targeted adversaries} to solve the adversarial model fairness problem.
 In our AT approach, we utilize PGD (targeted) attack \cite{madry2018towards} to find adversaries within the perturbation bound.
\autoref{eq_pgd_targeted} defines the PGD (targeted) attack, where $y_{\text{t}}$ defines the target labels:
\begin{align}
\label{eq_pgd_targeted}
x^{t+1} = \Pi_{\mathcal{B}(x, \epsilon)}\left(x^{t} + \alpha \cdot \text{sign}\left(\nabla_{x^t} L(f_{\theta}(x^t), y_{\text{t}})\right)\right).
\end{align}

Other additional methods, such as those proposed by ~\cite{moosavi-dezfooli_deepfool_2016,carlini_adversarial_2017}, are often used as probes to determine a model's accuracy under adversarial attack, referred to as the model's \emph{robust accuracy}.
Adversarial model fairness in classification is typically evaluated based on the class-wise robust accuracy performance of the classifier.
We also evaluate our approach using other standard adversarial attacks like AutoAttack, Squares \cite{croce_reliable_2020, andriushchenko_square_2020}.

%
\paragraph{Notation.} 
Following the notation in \cite{wei2023cfa}, we consider a function \( f_\theta:X \rightarrow Y \) parameterized by $\theta$ representing the K-class classification problem with \( Y = \{1, 2, \ldots, K\}\).
We consider an example \( x \) from \( X \) and respective label \(y\) from \(Y\), and perturb \( x \) within the $\ell_p$-norm perturbation bound \( B(x, \epsilon) = \{x' \mid \|x' - x\|_p \leq \epsilon\} \).
Then the clean and robust accuracy of the model $f_{\theta}$ are defined as \( A(f_\theta) \) and \( R(f_\theta) \):
\begin{align}
A(f_\theta) = \mathbb{E}_{(x, y) \sim D} [\chi(f_{\theta}(x) = y)],\qquad
\\    
R(f_\theta) = \mathbb{E}_{(x, y) \sim D} \left[\chi\left(\forall x' \in B(x, \epsilon), f_{\theta}(x') = y\right)\right],
\end{align}
where $\chi(\cdot)$ is the indicator function, equaling $1$ if the condition inside is true and $0$ otherwise.

\paragraph{Measuring Class-Wise Biases.}
Different metrics can be employed to assess the class biases of classification models.
A reliable aggregate measure is the \textbf{Class-Wise Accuracy ($C_{acc}$)}, computed for each class $c_j$ with $j\in \{1, \dots, K\}$, which has been predominantly used in previous works \cite{DBLP:conf/kdd/0001KJW021,Benz2020RobustnessMB}, and can be defined from samples of a validation set $\{ x_i\}_{i=1}^N$ of size $N$ with labels $y_i \in \{ c_j\}_{j=1}^K$  as:
\begin{align}
C_{acc}(c_j) = 
\frac{|\{x_i \mid f_{\theta}(x_i) = c_j, y_i = c_j \}|}{N}  + \nonumber \\
\frac{|\{x_i \mid f_{\theta}(x_i) \neq c_j, y_i \neq c_j \}|}{N},
\end{align}
where we consider all samples $x_i$ with label $y_i$ that are correctly classified as class $c_j$ by the model $f_{\theta}$ and all samples $x_i$ with label $y_i$ that are correctly classified as not being of class $c_j$.
Previous approaches such as \cite{DBLP:conf/kdd/0001KJW021,Benz2020RobustnessMB, wei2023cfa} predominantly use this metric to assess class-wise biases and further to calibrate adversarial training by scaling $\epsilon$ based on their class-wise robustness performance for each individual class during AT to achieve robust fairness \cite{wei2023cfa}.

As a further informative measure on class-wise biases, we consider the \textbf{Class False Positive Score ($C_{FPS}$)} to assess the confusion rates of each class $c_j$, i.e., how frequently samples from other classes are incorrectly classified as $c_j$. 
We use this measure to evaluate how the other group of classes are attracted to a specific class during AT.
While a high $C_{FPS}$ can also manifests in a decrease in class-wise accuracy, the imbalance it induces can hardly be addressed in an untargeted training setting.
It therefore measures the potential of leveraging targeted adversarial training for model fairness.
To calculate the $C_{FPS}$ for a specific class, we calculate the number of misclassifications where samples from other classes, i.e.~samples $x_i$ from the validation set $\{ x_i\}_{i=1}^N$ with labels $y_i \in \{ c_j\}_{j=1}^K$  are incorrectly classified by model $f_{\theta}$ as this particular class $c_j$, i.e. the cardinality of $\{x_i|f_{\theta}(x_i)=c_j, y_i \neq c_j \}$.
We then normalize this count by the total number of misclassifications across all classes:
\begin{align}
    C_{FPS}(c_j) = \frac{|\{x_i|f_{\theta} (x_i)=c_j, y_i \neq c_j \}|}{|\{x_i|f_{\theta} (x_i) \neq y_i \}|}.
\end{align}
\\
A higher $C_{FPS}$ for a class $c_j$ indicates that $c_j$ is frequently assigned by the network even for samples of different classes.
Regarding the latent space, one can hypothesize that a class with a high $C_{FPS}$ occupies a relatively large, under-determined region.

\section{Empirical Analysis on $C_{FPS}$}
\begin{figure}[htb]
\centering
\begin{subfigure}{0.46\textwidth}
    \centering
    \includegraphics[width=\linewidth]{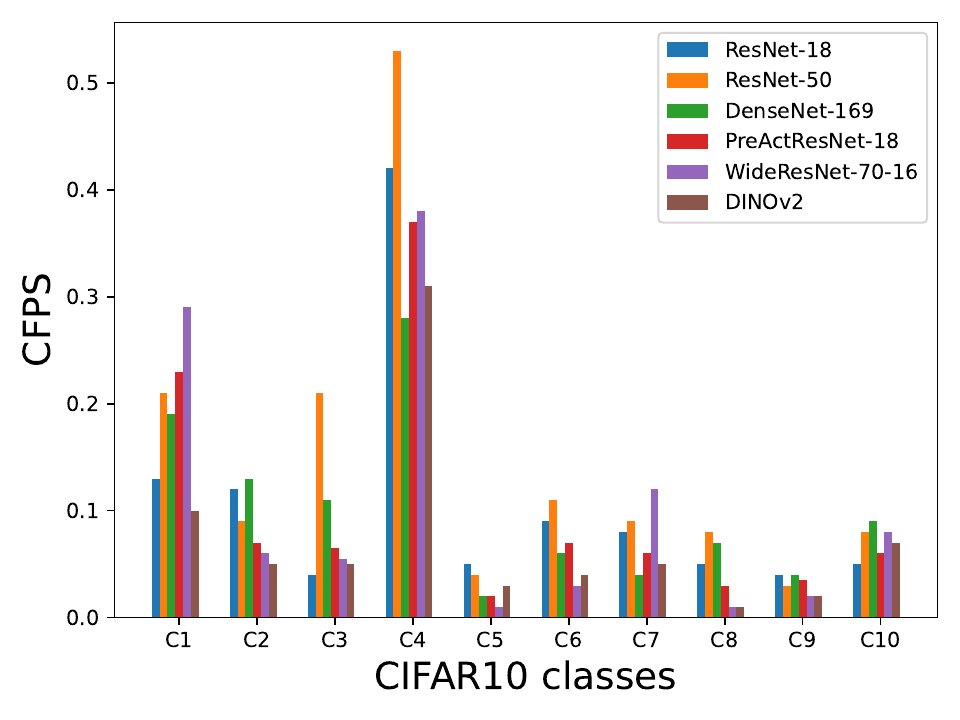}
\end{subfigure}
\caption{Class False Positive Scores ($C_{FPS}$) on the CIFAR-10 dataset using adversarially trained models.}
\label{fig:false_positive_scores}
\end{figure}

To better understand class vulnerability towards misclassifications, we analyze the $C_{FPS}$ to identify classes frequently misassigned by the model.
Utilizing this information, we aim to defend against perturbations leading to incorrect predictions of these classes by incorporating adversarial examples targeted towards these classes during AT.
\\
We conduct an experiment by evaluating $C_{FPS}$ on the CIFAR-10 dataset, employing robust models trained adversarially that are sourced from a standardized adversarial robustness benchmark \cite{croce2021robustbench, grabinski2022robust}.
\autoref{fig:false_positive_scores} illustrates the $C_{FPS}$ scores for CIFAR-10 classes.
A clear trend emerges in the $C_{FPS}$ scores across different adversarially trained architectures.
The horizontal axis in \autoref{fig:false_positive_scores} represent the 10 classes of CIFAR-10 in order.
Notably, class "cat", is frequently wrongly assigned to samples from other classes.
Similarly, class "airplane", although considered one of the easier classes due to its relatively high robust accuracy, also shows significant misclassification rates.
This indicates a need for a deeper understanding of the decision boundaries of these classes when perturbed.

To address the high confusion rates of specific classes, our work uses $C_{FPS}$ to model a prior multinomial distribution of classes.
This allows us to sample highly confused classes more frequently during \textit{targeted} adversarial training, where we employ targeted adversaries towards these classes to harden the model against class specific perturbations.

\section{FAIR-TAT}
\label{method}
\begin{figure}
   \centering   \includegraphics[width=\linewidth]{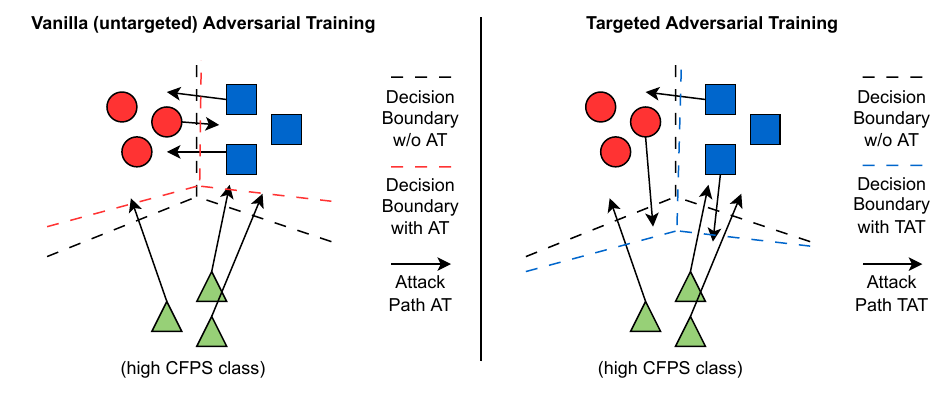}
   \caption{%
   We visualize the main idea of FAIR-TAT for a hypothetical three classes scenario in a two-dimensional latent space.
   Here, both classes red (circles) and blue (squares) are hard to distinguish from each other (and hence data points in the latent space are close to each other and the decision boundary), while green (triangles) is a distinctive class where data points are far from the decision boundary.
   When perturbed via an adversarial attack, data points move (and cross) the decision boundary along indicated paths (arrows).
   (\textbf{left}) During untargeted adversarial training, data points cross the decision boundary that is closest to them, which means no data point is perturbed towards the green class.
   Hence, the resulting new decision boundary (dashed red) is pushed away from the green class, which further improves its robustness and weakens the robustness of classes red and blue.
   (\textbf{right}) During targeted adversarial training, targets are sampled based on $C_{FPS}$ and therefore some red and blue data points are also targeted towards class green.
   This leads to an overall more fair decision boundary (dashed blue).%
   }
   \label{fig:fairtat}
\end{figure}

Based on the above empirical findings, prioritizing often confused classes during adversarial training can enhance the model's fairness by directing more adversarial samples towards these specific classes.
This allows to reduce target-specific confusion. 
Therefore, we incorporate targeted adversaries towards these classes more often during AT. 
When generating targeted adversaries, the choice of which targets to pick for training is crucial.
\textbf{Should we target all classes equally to promote a uniform learning across all classes during AT?}
Or \textbf{should we focus on specific target classes that are hard classes to reduce already existing class-wise imbalances?}
This decision shapes the fairness and effectiveness of the training process.
The optimal choice depends on the desired outcomes and data characteristics. In this work, we employ a score-based target sampling through evaluation of the false positive scores for individual classes for every instance during AT.
Our aim is thus to harden the worst performing classes, by allowing to better define their decision boundaries, not only when being used as source classes but also when defining the target class of an adversarial attack.
Please refer to \autoref{fig:fairtat} for a visualization of our main idea of FAIR-TAT.

\begin{algorithm}
\caption{FAIR-TAT}
\label{alg:Custom-TAT}
\scriptsize
\begin{algorithmic}[1]
\State {\textbf{Input:} A classifier $f_{\theta}(.)$; Training dataset $D$ with $N$ samples; Ground truth classes $Y$; Unique number of classes $K$, Prior distribution based on $C_{FPS}$ $\psi(0,K-1)$; Uniform distribution $\mu(0, K-1)$, where K is total number of classes; Batch size $B$; Weight average decay rate $\alpha$; Learning rate $\eta$; Training epochs S; Perturbation margin $\epsilon$; Adversarial loss $\ell$; Classification loss $L$; Hyper-parameter to scale base perturbation $\lambda_{1}$}

\State{\textbf{Output:} FAIR robust classifier $\Bar{f}_{\Bar{\theta}}(.)$}

\State{\texttt{/* Initialize parameters */}}
\State{$\theta \leftarrow \theta_{0}, \Bar{\theta} \leftarrow \theta$;}
\State{Split $D = D_{\text{train}} \cup D_{\text{valid}};$}

\For{$y \in Y$}
    \texttt{/* Initialize $\epsilon_{y}$ */} \\
    \texttt{$\epsilon_{y} \leftarrow \epsilon$;}
\EndFor

\For{\texttt{every epoch S $\leftarrow$ 1, 2, 3, $\ldots$,  N}}
    \For{\texttt{every minibatch ($x,y$) in $D_{\text{train}}$}}
        \If{\texttt{same based on $C_{FPS}$ prior}}
            \State \texttt{Sample targets:$y_{\text{t}} \sim \psi(0, K-1)$;}
        \Else 
            \State \texttt{Sample targets uniformly:$y_{\text{t}} \sim \mu(0, K-1)$;}
        \EndIf
        \If{\texttt{Any of $y_{\text{t}}$ == $y$}}
        \State \texttt{Resample targets again with respective target sampling}
        \EndIf
    %
        \State\texttt{Compute optimized targeted adversarial perturbation $\delta^*$}
        \State\texttt{$\delta^* = \arg\min_{\|\delta\| \leq \epsilon} \ell(f_\theta(x + \delta), y_{\text{target}})$;}
        \State \texttt{$x' = x + \delta^*$;}
        \State\texttt{$\theta \leftarrow \theta - \eta \Delta_{\theta}L_{\theta}(x', y)$;}
    \For{$y \in Y$ and $y_{t} \in Y$}
        \State \texttt{/* Update $R_{k}$, $C_{FPS-k}$ and $\epsilon_{k}$ */} 
        \State \texttt{$R_{k} \leftarrow $Train-Accuracy($f_{\theta},S$);}
        \State \texttt{$C_{FPS-k} \leftarrow $ $C_{FPS}$($f_{\theta},S$);}
        \State \texttt{$\epsilon_{k} \leftarrow (\lambda_{1} + r_{k})\epsilon$;}
    \EndFor
\EndFor
\EndFor
\State{\Return $\Bar{f}_{\Bar{\theta}}$;}
\end{algorithmic}
\end{algorithm}

\paragraph{Sampling of Targets.}

As discussed in \autoref{bias}, a high value of $C_{FPS}$ for a specific class indicates frequent misclassifications as other classes.
This vulnerability presents an opportunity to enhance model fairness by refining the decision boundary for this specific class.
By increasing the frequency of sampling for this class during training, we aim to improve the model's ability to correctly classify it, thereby reducing misclassifications towards it.

To achieve this, we evaluate $C_{FPS}$ for each instance during training and utilize these class-wise false positive scores as weights to model a prior multi-nominal distribution $\psi(0, K-1)$.
This $C_{FPS}$ based prior distribution guides the sampling of targets $y_{t} \sim \psi(0, K - 1)$ for the next training instance, where $K$ is number of unique classes.
Consequently, classes with high $C_{FPS}$ are sampled more often, indicating more weight on hard challenging classes throughout the training.
We ensure that sampled targets do not coincide with ground truth classes by setting their respective sampling probability to zero.
Research suggests that datasets like CIFAR-10 and CIFAR-100 contain classes that are challenging to distinguish due to closer semantic similarities among the classes.
By prioritizing these challenging or hard classes through sampling and by learning to defend against perturbations leading to misclassification towards these classes, we aim to improve the model's understanding of the semantic differences between the samples and improve fairness.
We have also considered uniform sampling of targets $y_{\text{t}} \sim \mu(0, K-1)$ to showcase the effectiveness of target sampling using prior based on $C_{FPS}$.
The results with respect to uniform sampling of targets are provided in the supplementary (see Section A.2).
\vspace{-0.1cm}
\paragraph{FAIR-TAT Framework.}
The overall FAIR-TAT framework is shown in \autoref{alg:Custom-TAT}.
Suppose for each adversarial training instance, assuming there are $y \in Y$ ground truth classes and $y_{t} \in Y$ target classes, the training robust accuracy of the specific $k$-th class during the last training interval is denoted by $R_{k}$.
Similarly, $C_{FPS-k}$ represents the class false positive score of the $k$-th class during the last training interval. We aim to leverage both these properties in our approach.
We utilize the class false positive scores to model a multinominal distribution which acts as prior for sampling the targets during our targeted adversarial scenario.
We use robust accuracy for customizing the perturbation margin for each class as showcased in \cite{wei2023cfa}, that various classes exhibit distinct preferences for perturbation margins during adversarial training, a critical factor in improving fairness.
Having smaller $\epsilon$ for hard classes than easy classes improves robustness and fairness of the model.
We use the update formula from \cite{wei2023cfa} for scaling the $\epsilon$, which is mentioned in line 27 of \autoref{alg:Custom-TAT}.
During inference, we use the same $\epsilon$ for every class to compare the results with other baselines that focus on fairness.
During the training phase, the customized perturbation margin $\epsilon_{k}$ converges to the optimal perturbation margin for each class.

\section{Experiments}
\label{sec:Experiments}

\begin{table*}[ht]
\centering
\caption{Comparison of PGD evaluations of FAIR-TAT framework with other methods focused on model fairness using PRN-18 on CIFAR-10 and CIFAR-100. Our method is marked with \ding{117}.}
\label{tab:comparison_1}
\scriptsize 
\begin{tabularx}{\textwidth}{l *{8}{>{\centering\arraybackslash}X}}
\toprule[2pt]
Method & \multicolumn{2}{c}{CIFAR-10} & \multicolumn{2}{c}{CIFAR-10} & \multicolumn{2}{c}{CIFAR-100} & \multicolumn{2}{c}{CIFAR-100} \\
& \multicolumn{2}{c}{(Clean Acc.)} & \multicolumn{2}{c}{(Robust Acc.)} & \multicolumn{2}{c}{(Clean Acc.)} & \multicolumn{2}{c}{(Robust Acc.)} \\
\cmidrule(r){2-3} \cmidrule(r){4-5} \cmidrule(r){6-7} \cmidrule(r){8-9}
 & Overall & Worst & Overall & Worst & Overall & Worst & Overall & Worst \\
\midrule[2pt]
AT & $83.5 $ & $67.3 $ & $\mathbf{47.9}$ & $19.6$ & $53.7$ & $21.0$ & $19.7$ & $1.6$ \\
FAIR-TAT \ding{117} & $ \mathbf{85.6}$ & $\mathbf{73.7}$ & $46.7$ & $\mathbf{19.7}$ & $\mathbf{61.4}$ & $\mathbf{25.0}$ & $\mathbf{20.9}$& $\mathbf{2.2}$ \\
\midrule
AT + EMA & $84.1$ & $67.2$ & $\mathbf{49.4} $ & $21.3$ & $57.7$ &$24.0$& $\mathbf{23.4}$ & $1.4$\\
FAIR-TAT + EMA \ding{117} & $\mathbf{85.8}$ & $\mathbf{72.4}$ & $47.9$ & $\mathbf{21.7}$ &$\mathbf{61.3}$  &$\mathbf{27.0}$ & $23.2$ &  $\mathbf{2.6}$\\
\midrule
 AT + CFA & $83.8$ & $68.1$ & $\mathbf{50.1}$ & $22.8$ & $59.3$ & $24.8$ & $24.7$ & $2.4$ \\
FAIR-TAT+CFA \ding{117} & $\mathbf{84.8}$ & $\mathbf{72.0}$ & $48.8$ & $\mathbf{24.6}$ & $\underline{\mathbf{62.8}}$ & $\underline{\mathbf{27.6}}$ & $\mathbf{25.0}$ & $\mathbf{2.9}$\\
\midrule
TRADES & $\mathbf{81.9}$ & $62.9$ & $\mathbf{53.1}$ & $ 24.3$ & $56.5$ & $13.0$ & $28.2$  & $2.0$ \\
FAIR-TAT + TRADES \ding{117} & $81.7 $ & $\mathbf{68.1}$ & $51.9 $ & $\mathbf{26.2}$ & $\mathbf{59.4}$ & $\mathbf{24.0}$ & $\mathbf{29.4}$ & $\mathbf{8.9}$\\
\midrule
TRADES + EMA & $\mathbf{82.6}$ & $65.3$ & $\underline{\mathbf{53.7}}$ & $24.6$ &$58.0$ & $17.9$ &$16.0$  &$3.0$  \\
FAIR-TAT + TRADES + EMA \ding{117} & $81.9$ & $ \mathbf{68.9}$ & $53.3$ & $\mathbf{26.8}$ & $\mathbf{59.4}$ &$\mathbf{26.0}$ & $\mathbf{27.7}$ & $\underline{\mathbf{9.1}}$ \\
\midrule
 TRADES + CFA & $\mathbf{82.3}$ & $65.4$ & $\underline{\mathbf{53.7}}$ & $25.2$ & $\mathbf{60.2}$ & $18.6$ & $29.4$ & $3.2$ \\
FAIR-TAT + TRADES + CFA \ding{117} & $82.1$ & $\mathbf{68.6}$ & $53.1$ & $\underline{\mathbf{26.9}}$ & $59.7$ & $\mathbf{26.4}$ & $\underline{\mathbf{29.6}}$ & $\mathbf{8.6}$ \\
\midrule
FAT & $84.9 $ & $68.5 $ & $\mathbf{48.1} $ & $18.7 $ &$54.0$  & $16.0$ & $19.3$ & $1.8$ \\
FAIR-TAT + FAT \ding{117} & $\mathbf{86.6} $ & $\underline{\mathbf{77.2}}$ & $46.1$ & $\mathbf{22.7}$ & $\mathbf{60.9}$ & $\mathbf{24.0}$ & $\mathbf{20.9}$ & $\mathbf{2.7}$ \\
\midrule
FAT + EMA & $86.3 $ & $69.2$ & $\mathbf{48.5} $ & $18.8 $ & $56.9$ & $22.0$ & $\mathbf{23.6}$ & $1.6$ \\
FAIR-TAT + FAT + EMA \ding{117} & $\underline{\mathbf{86.8}}$ & $\mathbf{73.5} $ & $47.5 $ & $\mathbf{21.3} $ &$\mathbf{61.6}$& $\mathbf{23.2}$ & $21.9$ & $\mathbf{2.8}$ \\
\midrule
FAT + CFA & $85.0$ & $71.0$ & $\mathbf{51.0}$ & $23.4$ & $59.8$ & $24.9$ & $\mathbf{24.1}$ & $2.0$\\
FAIR-TAT + FAT + CFA \ding{117} & $\mathbf{86.1}$ & $\mathbf{74.9}$ & $48.2$ & $\mathbf{24.4}$ & $\mathbf{60.0}$ &$\mathbf{25.3}$ & $23.1$ & $\mathbf{2.3}$ \\
\midrule[2pt]
FRL & $82.8 $ & $71.4 $ & $45.7 $ & $24.4 $ & $52.0$ & $23.0$ & $22.7$ & $1.2$ \\
FRL + EMA & $83.6 $ & $69.5 $ & $46.3$ & $24.8 $ & $54.0$ & $24.2$ &$22.5$  & $1.0$ \\
BAT + TRADES & $86.5$ & $73.4$ & $49.8$ & $22.1$ & $60.9$ &$25.9$& $24.4$ & $1.8$ \\
WAT + TRADES & $81.2$ & $65.9$ & $47.1$ & $26.3$ & $57.3$ &$21.2$& $25.6$ & $2.9$ \\
\midrule[2pt]
Clean Training & $94.0 $ & $79.4 $ & $13.7$ & $2.7$ & $76.7$ & $39.2$ & $4.1$ & $0$ \\
\bottomrule[2pt]
\end{tabularx}
\end{table*}
In this section, we demonstrate the effectiveness of our approach, FAIR-TAT, in enhancing the clean accuracy while maintaining the overall adversarial robustness, along with improving class-wise fairness towards clean and robust samples. Additionally, we evaluate the overall robustness and fairness of our method concerning common corruptions.
\\
\textbf{Data.}
Our experiments utilize the benchmark datasets CIFAR-10 and CIFAR-100 \cite{krizhevsky2009learning}.
We assess the performance of our approach under the following architectures: PRN-18, and XCiT-S12.
These architectures are widely recognized for assessing robustness \cite{croce2021robustbench}. In the main paper, we present results based on CIFAR-10 and CIFAR100 using the PRN-18 architecture.
Furthermore, we extend our evaluation to assess the model's robustness and fairness against common corruptions, utilizing the CIFAR-10C dataset \cite{hendrycks2019robustness}.
Notably, this study marks the first evaluation of common corruptions within the context of model fairness towards classification.
\\
\textbf{Metrics and Evaluations.}
To demonstrate the robustness and fairness of the model, we report both the overall average clean and robust accuracy, as well as the clean and robust accuracy of the hard class in the form of worst class accuracy.
\\
\textbf{Baselines.}
\label{baselines}
For our primary baseline, we select Vanilla Adversarial Training (AT) \cite{madry2018towards}, TRADES \cite{zhang2019theoretically} and FAT \cite{zhang2020attacks}.
Given that the primary objective of our model is fairness, we compare our approach to FRL \cite{xu2021robust}, which is the first adversarial training algorithm specifically designed to improve class-wise robustness and fairness.
Additionally, we include CFA \cite{wei2023cfa}, a recent instance-based adaptive adversarial training approach that dynamically customizes perturbation margins during adversarial training. 
\begin{table*}[ht]
\centering
\caption{Comparison of AutoAttack and Squares evaluations of FAIR-TAT framework with other methods focused on model fairness using PRN-18 on CIFAR-10 and CIFAR-100. Our method is marked with \ding{117}.}
\label{tab:comparison_2}
\scriptsize 
\begin{tabularx}{\textwidth}{l *{6}{>{\centering\arraybackslash}X}}
\toprule[2pt]
Method & \multicolumn{2}{c}{CIFAR-10 (AA)} & \multicolumn{2}{c}{CIFAR-10 (Squares)} & \multicolumn{2}{c}  {CIFAR-100 (AA)} \\
\cmidrule(r){2-3} \cmidrule(r){4-5} \cmidrule(r){6-7}
 & Overall & Worst Class & Overall & Worst Class & Overall & Worst Class \\
\midrule[2pt]
 AT  & $\mathbf{45.7}$ & $15.4$ & $51.3$ & $19.4$ & $17.8$ & $1.3$ \\
FAIR-TAT \ding{117} & $45.0$ & $\mathbf{18.7}$ & $\mathbf{51.8}$ & $\mathbf{26.7}$ & $\mathbf{19.5}$ & $\mathbf{1.9}$ \\
\midrule
AT + EMA & $\mathbf{45.6}$ & $15.4$ & $51.7$ & $19.7$ & $\mathbf{21.6}$ & $1.2$ \\
FAIR-TAT + EMA \ding{117} & $45.0$ & $\mathbf{18.8}$ & $\mathbf{51.8}$ & $\mathbf{26.7}$ & $21.3$ & $\mathbf{2.0}$ \\
\midrule
 AT + CFA & $\mathbf{47.4}$ & $19.3$ & $51.7$ & $19.9$ & $21.9$ & $1.8$ \\
FAIR-TAT+CFA \ding{117} & $47.0$ & $\mathbf{24.1}$ & $\mathbf{52.6}$ & $\underline{\mathbf{31.8}}$ & $\mathbf{22.3}$ & $\mathbf{2.1}$ \\
\midrule
 TRADES & $\mathbf{49.8}$ & $18.7$ & $\mathbf{54.0}$ & $22.4$ & $25.9$ & $1.7$ \\
FAIR-TAT + TRADES \ding{117} & $48.6$ & $\mathbf{20.8}$ & $52.6$ & $\mathbf{26.1}$ & $\underline{\mathbf{26.7}}$ & $\underline{\mathbf{4.8}}$\\
\midrule
TRADES + EMA & $\mathbf{49.8}$ & $18.6$ & $\mathbf{54.0}$ & $22.4$ & $14.9$ & $2.1$ \\
FAIR-TAT + TRADES + EMA \ding{117} & $48.6$ & $\mathbf{21.1}$ & ${52.6}$ & $\mathbf{26.1}$ & $\mathbf{24.9}$ & $\mathbf{4.0}$ \\
 \midrule
TRADES + CFA                       & $\underline{\mathbf{50.3}}$ & $21.3$          & $\underline{\mathbf{54.2}}$ & $22.6$          & $\mathbf{26.2}$ & $2.9$ \\
FAIR-TAT + TRADES + CFA \ding{117} & $49.8$          & $\mathbf{24.0}$ & $53.0$          & $\mathbf{25.8}$ & $24.7$          & $\mathbf{4.7}$ \\
\midrule
FAT                       & $44.2$          & $16.0$          & $50.8$          & $20.8$          & $17.6$          & $1.3$\\
FAIR-TAT + FAT \ding{117} & $\mathbf{44.6}$ & $\mathbf{17.9}$ & $\mathbf{51.5}$ & $\mathbf{26.3}$ & $\mathbf{18.4}$ & $\mathbf{1.8}$\\
\midrule
FAT + EMA                      & $44.2$         & $15.9$         & $50.8$         & $20.8$         & $\mathbf{20.3}$ & $1.3$ \\
FAIR-TAT + FAT+ EMA \ding{117} & $\mathbf{44.6}$ & $\mathbf{18.1}$ & $\mathbf{51.5}$ & $\mathbf{26.3}$ & $18.6$         & $\mathbf{2.0}$ \\
\midrule
FAT + CFA                       & $\mathbf{49.4}$ & $\mathbf{22.6}$ & $\mathbf{53.4}$ & $24.1$          & $\mathbf{21.3}$ & $\mathbf{1.7}$ \\
FAIR-TAT + FAT + CFA \ding{117} & $44.3$          & $22.1$          & $49.0$          & $\mathbf{24.4}$ & $20.2$          & $1.5$\\
\midrule[2pt]
FRL & $44.0$ & $23.2$ & $49.7$ & $24.6$ & $20.1$ & $1.2$ \\
FRL + EMA & $44.2$ & $23.9$ & $50.8$ & $24.9$ & $19.9$ & $1.0$ \\
BAT + TRADES & $44.1$ & $18.7$ & $51.5$ &$25.0$& $21.0$ & $1.0$ \\
WAT + TRADES & $45.9$ & $\underline{24.3}$ & $51.7$ &$26.2$& $22.6$ & $2.1$ \\
\bottomrule[2pt]
\end{tabularx}

\end{table*}
We also adapt CFA \cite{wei2023cfa}, TRADES \cite{zhang2019theoretically} and FAT \cite{zhang2020attacks} according to our approach and compare the results with their respective baseline versions.
We include recent fairness baselines such as  BAT \cite{sun2023improving} and WAT \cite{li2023wat}.
Furthermore, we apply weight averaging techniques, such as Fairness Aware Weight Average (FAWA) \cite{wei2023cfa} as in CFA and Exponential Moving Average (EMA) as well for our approach. Note that we have reported the results from the last checkpoint evaluations for all the methods therefore the results varies when reported with best checkpoint version.
Also, The worst class accuracy for CIFAR-100 is defined as the average performance of the worst 10\% of classes following the baseline works.
\\
\textbf{Training Settings.} 
We train PRN-18 network in accordance with the settings outlined in \cite{rice2020overfitting}.
Both networks undergo training using stochastic gradient descent (SGD) with a momentum of $0.9$ and weight decay set at $5 \times 10^{-4}$.
The learning rate scheduler is initially configured at $0.1$.
Learning rate adjustments are made by dividing the rate by $10$ after 50\% and 75\% of the epochs during training.
Adversarial training is conducted with a default perturbation margin of $\epsilon = 8/255$ and a robustness regularization parameter of $\beta = 2$.
The parameter $\lambda_{1}$ is set to $0.5$, which is used for scaling the customized perturbation of the classes.
To enhance the initialization process, the weight averaging procedure begins after the 50th epoch.
For the FAWA weight averaging method, a validation set is constructed by utilizing 2\% of samples from each class, while the remaining 98\% of train set is employed for training.
This approach ensures that FAWA does not impose a substantial computational burden.
The fairness threshold for FAWA is established at $0.2$. All the other baseline versions are trained using their respective training configurations. We conducted our experiments on internal cluster with 8 RTX A1000 GPUs.
\\
\textbf{Performance of Fairness and Robustness on PGD. }
We compare our approach, FAIR-TAT, with other methods aimed at improving robust model fairness.
All baseline methods mentioned in \autoref{sec:Experiments}, including ours, are trained using PGD. \autoref{tab:comparison_1} presents the clean and robust accuracy results for the CIFAR-10 and CIFAR-100 dataset using the PRN-18 architecture, with robustness evaluated based on PGD attacks.
In \autoref{tab:comparison_1}, "AT" refers to conventional adversarial training, while "FAIR-TAT" is our comparitive adversarial approach.
We also show adaptability of our approach to other methods for comparison, such as comparing TRADES with FAIR-TAT + TRADES.
To be fair, each approach customizes perturbation margins for individual class during training.
Results without customized perturbation margins are provided in the supplementary material (see section A.3). 

From \autoref{tab:comparison_1}, we observe that FAIR-TAT significantly improves clean accuracy, both overall and for the worst class, in both CIFAR-10 and CIFAR-100 datasets.
Additionally, it enhances worst-class robust accuracy while maintaining nearly the same overall robustness, with only marginal reductions in case of CIFAR-10, and improves overall robustness in CIFAR-100.
This improvement in CIFAR-100 can be attributed to the dataset's diversity and larger number of classes, which enhances the ability to differentiate features more effectively.
Overall, these results suggest that FAIR-TAT offers a better trade-off between robustness and fairness. 

We also compare FAIR-TAT with weight averaging techniques such as EMA and FAWA, as used in CFA \cite{wei2023cfa}.
Our method maintains fairness after applying these techniques.
Specifically, for the larger number of classes in CIFAR-100, FAIR-TAT with TRADES setting outperforms other baselines in both fairness and robustness, demonstrating that our method scales well with an increasing number of classes.
Overall, our FAIR-TAT approach outperforms state-of-the-art methods in terms of clean accuracy and fairness when evaluated against the adversaries it was trained on, though with a minor trade-off in robust accuracy in certain cases.

\begin{table}[ht!]
\centering
\tiny
\caption{\small{Minimum Class accuracies of FAIR-TAT method on common corruptions. T: TRADES, F: FAT}}
\label{tab:fairtat-corr}
\begin{tabularx}{\columnwidth}{@{}l*{9}{@{}X@{}}}
\toprule
\makecell{Corruption Type \\ (C)} & \makecell{FAIRTAT} & \makecell{FAIRTAT \\ (EMA)} & \makecell{FAIRTAT \\ (CFA)} & \makecell{FAIRTAT \\ (T)} & \makecell{FAIRTAT \\ (T+EMA)} & \makecell{FAIRTAT \\ (T+CFA)} & \makecell{FAIRTAT \\ (F)} & \makecell{FAIRTAT \\ (F+EMA)} & \makecell{FAIRTAT \\ (F+CFA)} \\
\midrule
gaussian\_noise & \num{0.63} & \num{0.61} & \num{0.58} & \num{0.60} & \num{0.60} & \num{0.58} & \num{0.58} & \num{0.63} & \underline{\num{0.66}} \\
shot\_noise & \num{0.66} & \num{0.64} & \num{0.60} & \num{0.61} & \num{0.61} & \num{0.61} & \num{0.61} & \num{0.66} & \underline{\num{0.68}} \\
speckle\_noise & \num{0.66} & \num{0.65} & \num{0.62} & \num{0.61} & \num{0.61} & \num{0.61} & \num{0.61} & \num{0.66} & \underline{\num{0.67}} \\
impulse\_noise & \num{0.51} & \num{0.49} & \num{0.52} & \num{0.39} & \num{0.46} & \num{0.66} & \num{0.66} & \underline{\num{0.68}} & \num{0.64} \\
defocus\_blur & \num{0.68} & \num{0.67} & \num{0.66} & \num{0.70} & \num{0.68} & \underline{\num{0.71}} & \underline{\num{0.71}} & \num{0.68} & \num{0.69} \\
gaussian\_blur & \num{0.65} & \num{0.67} & \num{0.65} & \num{0.64} & \num{0.62} & \underline{\num{0.68}} & \num{0.62} & \num{0.62} & \num{0.37} \\
motion\_blur & \num{0.59} & \num{0.62} & \num{0.58} & \num{0.62} & \num{0.61} & \num{0.64} & \num{0.61} & \underline{\num{0.67}} & \num{0.45} \\
zoom\_blur & \num{0.67} & \underline{\num{0.70}} & \num{0.64} & \num{0.65} & \num{0.63} & \num{0.64} & \num{0.64} & \num{0.64} & \num{0.40} \\
snow & \num{0.63} & \num{0.63} & \num{0.63} & \num{0.58} & \num{0.55} & \num{0.55} & \underline{\num{0.64}} & \underline{\num{0.64}} & \num{0.63} \\
fog & \num{0.45} & \num{0.39} & \num{0.34} & \underline{\num{0.47}} & \num{0.20} & \num{0.17} & \num{0.45} & \num{0.32} & \num{0.15} \\
brightness & \num{0.63} & \underline{\num{0.72}} & \num{0.65} & \num{0.65} & \num{0.63} & \num{0.66} & \underline{\num{0.72}} & \num{0.67} & \underline{\num{0.72}} \\
contrast & \num{0.22} & \num{0.20} & \num{0.17} & \num{0.22} & \num{0.20} & \num{0.18} & \underline{\num{0.24}} & \num{0.18} & \num{0.17} \\
elastic\_transform & \num{0.66} & \underline{\num{0.67}} & \num{0.61} & \num{0.62} & \num{0.64} & \num{0.64} & \num{0.65} & \num{0.64} & \num{0.65} \\
pixelate & \num{0.66} & \num{0.68} & \num{0.63} & \num{0.64} & \num{0.66} & \num{0.63} & \num{0.68} & \num{0.68} & \underline{\num{0.73}} \\
jpeg\_compression & \num{0.66} & \underline{\num{0.71}} & \num{0.63} & \num{0.64} & \num{0.63} & \num{0.67} & \num{0.68} & \num{0.68} & \underline{\num{0.71}} \\
spatter & \underline{\num{0.64}} & \underline{\num{0.64}} & \num{0.60} & \num{0.62} & \num{0.62} & \num{0.62} & \num{0.62} & \num{0.63} & \num{0.63} \\
saturate & \underline{\num{0.67}} & \num{0.66} & \num{0.61} & \num{0.63} & \num{0.63} & \num{0.66} & \num{0.64} & \num{0.66} & \num{0.66} \\
frost & \num{0.60} & \num{0.57} & \num{0.54} & \num{0.58} & \num{0.54} & \num{0.58} & \num{0.62} & \num{0.62} & \underline{\num{0.63}} \\
\bottomrule
\end{tabularx}
\end{table}
\paragraph{\textbf{Transferability to Other Adversaries.}}
To evaluate the transferability of our approach, we extend our evaluation to adversaries that were not initially trained on.
We use AutoAttack \cite{croce2020reliable}, a widely used ensemble adversarial attack and Squares attack \cite{andriushchenko_square_2020}, a black-box attack to evaluate the adversarial robustness to access our approach compared to baselines.
We utilize the same baselines as in \autoref{tab:comparison_1}.
\autoref{tab:comparison_2} showcases the evaluations of AutoAttack and Squares on CIFAR-10 and AutoAttack on CIFAR-100 using PRN-18.
Standard AutoAttack itself is an ensemble that contains Squares attack, however, we present the Squares evaluation separately to show the performance of our method when evaluated on a black-box attack alone.
From the evaluations, it is clear that our approach, FAIR-TAT with respective weight averaging schemes, provides better transferability of fairness by enhancing the worst class accuracy while maintaining the marginal overall model robustness when assessed against different adversaries outside of the training domain.

\begin{table}[ht!] 
\centering
\tiny
\caption{\small{Minimum Class accuracies of baseline methods on common corruptions. T: TRADES, F: FAT}}
\label{tab:baselines-corrup}
\begin{tabularx}{\columnwidth}{@{}l*{9}{@{}X@{}}}
\toprule
\makecell{Corruption Type \\ (C)} & \makecell{AT} & \makecell{AT \\ (EMA)} & \makecell{AT \\ (CFA)} & \makecell{AT \\ (T)} & \makecell{AT \\ (T+EMA)} & \makecell{AT \\ (T+CFA)} & \makecell{AT \\ (F)} & \makecell{AT \\ (F+EMA)} & \makecell{AT \\ (F+CFA)} \\ 
\midrule
gaussian\_noise & \num{0.57} & \num{0.58} & \num{0.54} & \num{0.52} & \num{0.55} & \num{0.54} & \underline{\num{0.66}} & \num{0.63} & \num{0.44} \\ 
shot\_noise & \num{0.58} & \num{0.60} & \num{0.56} & \num{0.53} & \num{0.57} & \num{0.56} & \underline{\num{0.68}} & \num{0.66} & \num{0.45} \\ 
speckle\_noise & \num{0.58} & \num{0.60} & \num{0.56} & \num{0.54} & \num{0.57} & \num{0.56} & \underline{\num{0.67}} & \num{0.66} & \num{0.44} \\ 
impulse\_noise & \num{0.52} & \num{0.53} & \num{0.49} & \num{0.46} & \num{0.51} & \num{0.52} & \num{0.64} & \num{0.66} & \underline{\num{0.68}} \\ 
defocus\_blur & \num{0.64} & \num{0.66} & \num{0.70} & \num{0.62} & \num{0.63} & \num{0.64} & \underline{\num{0.71}} & \num{0.68} & \num{0.69} \\ 
gaussian\_blur & \num{0.63} & \num{0.66} & \underline{\num{0.68}} & \num{0.59} & \num{0.64} & \num{0.59} & \num{0.62} & \num{0.62} & \num{0.40} \\ 
motion\_blur & \num{0.62} & \num{0.62} & \underline{\num{0.63}} & \num{0.57} & \num{0.62} & \num{0.58} & \num{0.62} & \num{0.61} & \num{0.45} \\ 
zoom\_blur & \num{0.64} & \num{0.67} & \underline{\num{0.68}} & \num{0.59} & \num{0.65} & \num{0.59} & \num{0.64} & \num{0.68} & \num{0.40} \\ 
snow & \num{0.53} & \num{0.59} & \num{0.55} & \num{0.56} & \num{0.58} & \underline{\num{0.64}} & \num{0.63} & \num{0.63} & \num{0.63} \\ 
fog & \num{0.46} & \num{0.43} & \num{0.43} & \num{0.42} & \underline{\num{0.47}} & \num{0.34} & \num{0.45} & \num{0.32} & \num{0.15} \\ 
brightness & \num{0.62} & \num{0.67} & \underline{\num{0.72}} & \num{0.63} & \num{0.65} & \underline{\num{0.72}} & \num{0.67} & \num{0.71} & \underline{\num{0.72}} \\ 
contrast & \underline{\num{0.22}} & \num{0.17} & \num{0.17} & \num{0.24} & \num{0.18} & \num{0.18} & \num{0.18} & \num{0.17} & \underline{\num{0.22}} \\ 
elastic\_transform & \underline{\num{0.66}} & \num{0.61} & \num{0.64} & \num{0.65} & \num{0.62} & \num{0.64} & \num{0.64} & \num{0.65} & \num{0.65} \\ 
pixelate & \num{0.66} & \num{0.63} & \num{0.68} & \num{0.68} & \num{0.68} & \num{0.63} & \num{0.68} & \underline{\num{0.73}} & \num{0.64} \\ 
jpeg\_compression & \num{0.66} & \num{0.63} & \num{0.67} & \num{0.68} & \num{0.68} & \num{0.67} & \num{0.68} & \underline{\num{0.71}} & \num{0.64} \\ 
spatter & \underline{\num{0.64}} & \num{0.60} & \num{0.62} & \num{0.62} & \num{0.62} & \num{0.62} & \num{0.63} & \num{0.63} & \num{0.62} \\ 
saturate & \underline{\num{0.67}} & \num{0.61} & \num{0.66} & \num{0.66} & \num{0.63} & \num{0.66} & \num{0.66} & \num{0.66} & \num{0.66} \\ 
frost & \num{0.60} & \num{0.54} & \num{0.58} & \num{0.62} & \num{0.62} & \num{0.62} & \underline{\num{0.63}} & \num{0.62} & \num{0.62} \\ 
\bottomrule 
\end{tabularx} 
\end{table}
\vspace{-0.2cm}
\paragraph{Fairness and Robustness on Common Corruptions.}
In this work, we evaluate the robustness and fairness of models trained using our approach on common corruptions compared to baseline methods.
We use the CIFAR-10C dataset, which contains 18 types of corruptions applied to CIFAR-10 images.
\autoref{tab:fairtat-corr} and \autoref{tab:baselines-corrup} present the worst class accuracies for our method and the baselines, respectively.
\autoref{tab:fairtat-corr} shows the minimum class accuracies of FAIR-TAT adapted to other approaches, while \autoref{tab:baselines-corrup} lists the minimum class accuracies of the baselines. The best results on common corruptions of baselines and our methods are underlined in the tables. Our approach FAIR-TAT, when adapted to baselines, outperforms them in terms of worst class accuracies and overall model performance.
The overall accuracies on common corruptions are provided in the supplementary (see section B.3).
Overall, our findings suggest that FAIR-TAT improves both robustness and fairness compared to existing methods, with rare exceptions.
Although FAT performs comparably equally well on common corruptions, our approach achieves a better balance in performance considering fairness on corruptions as well on adversarial samples.
\vspace{-0.1cm}
\paragraph{Discussion of Limitations.}
FAIR-TAT enhances worst-class accuracy, but there's still significant room to improve class-wise balance.
We see our method as a step toward making models both robust and fair.
While FAIR-TAT offers slightly better trade-offs between clean and robust accuracy than traditional AT, clean accuracy still declines.
Technically, further exploration is needed; for example, we currently sample targets based on class-wise misclassification using $C_{FPS}$.
Although focusing on sample-wise misclassifications instead of class-wise misclassifications would be more beneficial, it is costly, and we plan to address this in future work.

\vspace{-0.2cm}
\paragraph{Broader Impact.}
Ensuring both robustness and fairness in neural models is important for their deployment in safety-critical applications.
Our work focuses in the direction of training a robust and fair classifier.
Robust and fair models maintain reliability under diverse conditions, reducing the risk of unexpected failures.
Fair models ensure equitable treatment across all demographics, preventing bias-related harm.
Although improvements in robustness of classifiers are desirable, they can also lead to a false sense of security. 
\section{Conclusion}
In conclusion, our empirical analysis highlights the need to focus more on specific classes which are prone to be misclassified often than others.
Building upon these findings, we introduce the Fair Targeted Adversarial Training (FAIR-TAT) framework which samples these most confused classes more often than others and learn to defend these class-specific perturbations by generating targeted adversaries.
This allows to learn the class-to-class bias more effectively.
FAIR-TAT demonstrates improved fairness against adversarial attacks compared to state-of-the-art baseline models not only on the attacks it is trained on but also on other type of adversaries.
Simultaneously, FAIR-TAT enhances accuracy on clean samples and common corruptions, outperforming existing baselines showcasing the overall effectiveness and balance of our approach against diverse threats.
\section*{Acknowledgements}
Steffen Jung and Margret Keuper acknowledge funding by the DFG Research
Unit 5336 - Learning to Sense.
\newpage
{\small
\bibliographystyle{ieee_fullname}
\bibliography{egbib}
}
\end{document}